\documentclass[10pt,twocolumn,letterpaper]{article}
\usepackage{iccv}
\usepackage{times}
\usepackage{epsfig}
\usepackage{graphicx}
\usepackage{amsmath}
\usepackage{amssymb}
\usepackage{color}
\usepackage{bm}
\usepackage{multirow}
\usepackage{diagbox}
\usepackage{array}
\usepackage{booktabs}
\usepackage{nth}
\usepackage{nicefrac}
\usepackage{relsize}
\usepackage{xspace} 
\usepackage{bm}
\usepackage{enumitem}
\usepackage{multirow}
\usepackage{paralist}

\makeatletter
\newcommand{\thickhline}{%
    \noalign {\ifnum 0=`}\fi \hrule height 1pt
    \futurelet \reserved@a \@xhline
}

\usepackage{cuted}
\usepackage{capt-of}
\newcolumntype{P}[1]{>{\centering\arraybackslash}p{#1}}
\newcommand{\mytilde}{\raise.17ex\hbox{$\scriptstyle\mathtt{\sim}$}}

\usepackage[pagebackref=true,breaklinks=true,colorlinks,bookmarks=false]{hyperref}

\usepackage[capitalize]{cleveref}
\crefname{section}{Sec.}{Secs.}
\Crefname{section}{Section}{Sections}
\Crefname{table}{Table}{Tables}
\crefname{table}{Tab.}{Tabs.}

\usepackage[breaklinks=true,bookmarks=false]{hyperref}

\iccvfinalcopy 

\def\iccvPaperID{11393} 

\ificcvfinal\fi

\begin{document}

\title{Neural Rendering of Humans in Novel View and Pose from Monocular Video}
\author{Tiantian Wang$^{1*}$, Nikolaos Sarafianos$^2$, Ming-Hsuan Yang$^1$, Tony Tung$^2$\\
$^{1}$University of California, Merced, $^{2}$Meta Reality Labs Research, Sausalito\\
}
\maketitle

\begin{abstract}
We introduce a new method that generates photo-realistic humans under novel views and poses given a monocular video as input. Despite the significant progress recently on this topic, with several methods exploring shared canonical neural radiance fields in dynamic scene scenarios, learning a user-controlled model for unseen poses remains a challenging task. To tackle this problem, we introduce an effective method to a) integrate observations across several frames and b) encode the appearance at each individual frame. We accomplish this by utilizing both the human pose that models the body shape as well as point clouds that partially cover the human as input. Our approach simultaneously learns a shared set of latent codes anchored to the human pose among several frames, and an appearance-dependent code anchored to incomplete point clouds generated by each frame and its predicted depth. The former human pose-based code models the shape of the performer whereas the latter point cloud-based code predicts fine-level details and reasons about missing structures at the unseen poses. To further recover non-visible regions in query frames, we employ a temporal transformer to integrate features of points in query frames and tracked body points from automatically-selected key frames. Experiments on various sequences of dynamic humans from different datasets including ZJU-MoCap show that our method significantly outperforms existing approaches under unseen poses and novel views given monocular videos as input. 
\end{abstract}

\begin{figure}[t]
    \centering
    \includegraphics[width=\linewidth]{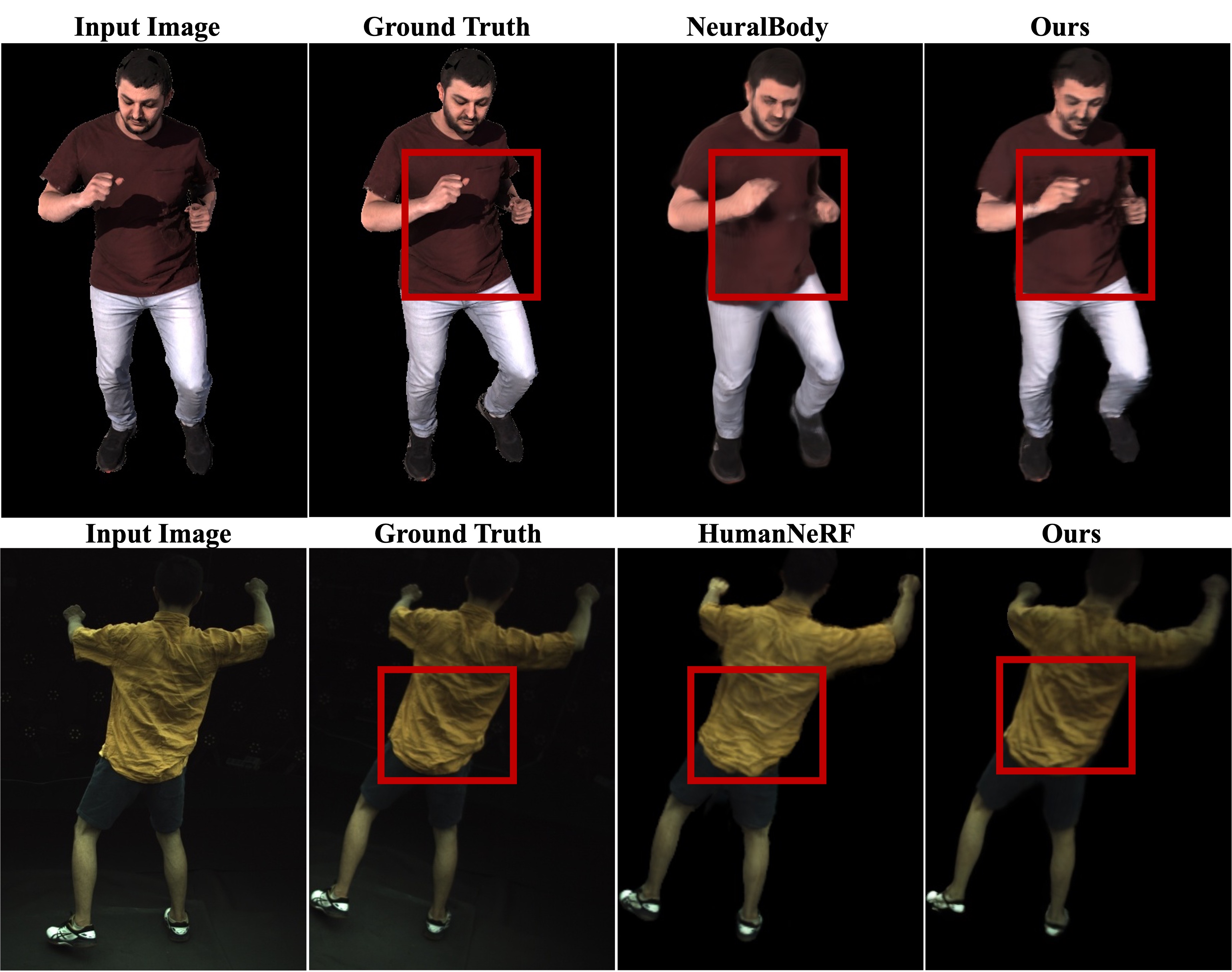}
    \caption{\textbf{Novel View Synthesis on Unseen Poses}. Given a monocular video, we predict novel views with body poses unseen from training with fine-level details (wrinkles) that works such as NeuralBody~\cite{peng2021animatable} or HumanNeRF~\cite{weng2022humannerf} struggle to obtain.}
    \label{fig:intro}
\vspace{-4mm}
\end{figure}

\section{Introduction}\label{sec:intro}

We set out to develop a method that generates photo-realistic humans under novel views and unseen poses from monocular RGB videos. 
To represent static scenes, neural radiance fields (NeRF)~\cite{mildenhall2020nerf} learn an implicit representation using neural networks, which has enabled photo-realistic rendering of shape and appearance from images. 
With dense multi-view observations as input, NeRF encodes density and color as a function of 3D coordinates and viewing directions by the MLPs along with a differentiable renderer to synthesize novel views. 
While it shows unprecedented visual quality on static scenes, applying it to high quality free-viewpoint rendering of humans in dynamic videos remains a challenging task.
Aiming to generalize NeRF to dynamic videos, D-NeRF~\cite{pumarola2020d} encodes a time step $t$ to differentiate motions across frames and converts scenes from the observation space to a shared canonical space to model the neural radiance field. 
As such, they can handle dynamic scenes to some extent but the poses remain uncontrollable by users. 
Furthermore, some approaches~\cite{yariv2020multiview,peng2020neural} introduce human pose as an additional input to serve as a geometric guidance for different frames. However, they either cannot generalize to novel poses or need more than one input view.

To overcome these limitations, we propose a novel approach by learning implicit radiance fields based on pose and appearance representations for high fidelity novel view and pose synthesis.
We leverage the human pose extracted from the parametric body model as a geometric prior to model motion information across frames. Shared latent codes anchored to the human poses are optimized, so that they integrate information across frames. 
However, a model that only formulates latent codes in a shared space will not generalize well to unseen poses without test-time optimization of the latent codes. 
To address this, we propose to model the appearance information by utilizing single-view depth information obtained by a depth estimation network.
Our model learns the appearance code anchored to incomplete point clouds in the 3D space. 
Point clouds are obtained by using single-view depth information to lift the RGB image to the 3D space, which provides partial information of the visible parts of the human body. 
The learned implicit representation enables reasoning of the unknown regions and complements the missing details on the human body.

To further leverage the temporal information from multiple frames, we introduce a temporal transformer that aggregates the trackable information. 
We utilize the parametric body model to track points from the query frame to the key frames. 
Following that, based on the learned implicit representation, we extract the pose code across frames and feed it into the temporal transformer for feature aggregation. 
Our method is extensively evaluated against state-of-the-art techniques on several sequences of humans in motion and exhibits significantly higher rendering quality under new views and unseen poses. 
In addition, we reconstruct fine-level details such as cloth wrinkles, hand details at a resolution and fidelity that several prior top-performing methods such as NeuralBody~\cite{peng2020neural} or HumanNeRF~\cite{weng2022humannerf} fail to recover (Figure~\ref{fig:intro}). 
The contributions of this work are:
\begin{compactitem}
    \item A new novel view synthesis framework that shows significant improvement on unseen poses compared to existing video methods, with high-fidelity reconstruction of fine-level facial, cloth and body details.
    \item We combine pose and appearance representations by modeling shared information across frames and specific information at each individual frame. 
    These two representations help generalize better to novel poses compared to only utilizing the pose representation.
    \item A temporal transformer is introduced to combine information across frames, which helps to recover non-visible details in the query frame (at unseen views).
\end{compactitem}

\section{Related Work}
\label{sec:relatedwork}
\noindent \textbf{3D Neural Representations}. 
Early 3D shape representation works can be classified into three categories: point-based methods~\cite{achlioptas2018learning,qi2017pointnet}, voxel-based methods~\cite{choy20163d, wu20153d} and mesh-based methods~\cite{ben2018multi, groueix2018papier,yang2017foldingnet}.
\begin{figure*}[t]
\centering
\includegraphics[width=0.98\linewidth]{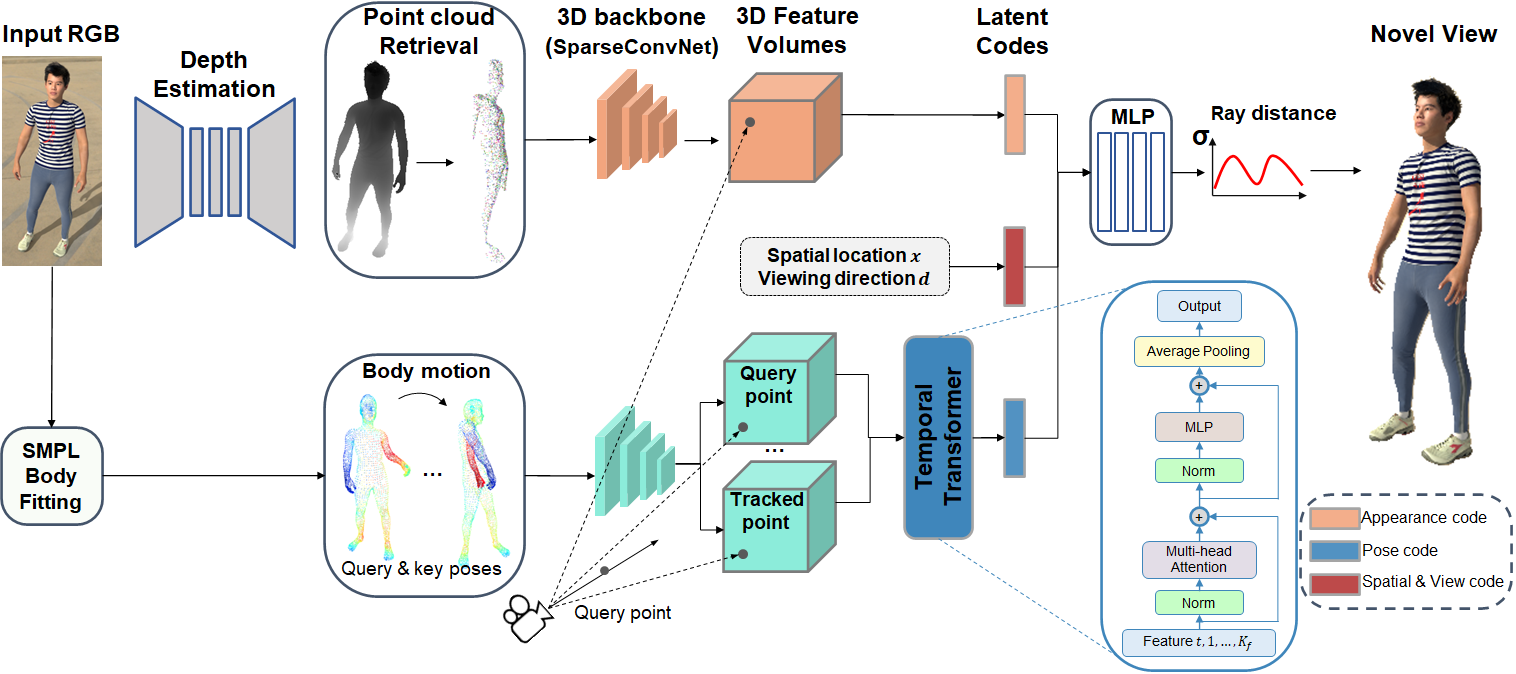}
\caption{\textbf{Overview.} Given a query point as input, our method learns pose and appearance codes which can simultaneously integrate shared information across frames and model the appearance information at each frame. The pose and appearance codes are anchored to the human pose and point clouds, respectively. The points clouds are obtained by lifting 2D information to the 3D using the predicted depth. In addition, we use the body motion to track 3D points from the query frame to the key frames and extract the pose code from the key frames. Given the pose codes from the query point and tracked points as input, a temporal transformer aggregates the codes. We use spatial location and viewing direction as extra inputs and train the model to predict the density and color for each 3D point. }
\label{fig:pipeline}
\vspace{-5mm}
\end{figure*}
Implicit representations are then used to represent shapes by reconstructing a continuous surface geometry, which utilizes the spatial coordinates as the input and outputs the signed distances or occupancy values. 
With advances in differential rendering methods, geometry and appearance can be learned from multi-view observations. 
Related works can be categorized into static~\cite{liu2020nsvf,mildenhall2020nerf,sitzmann2019deepvoxels,sitzmann2019scene,yu2021pixelnerf,yu2021plenoctrees} and dynamic scenes~\cite{chen2021animatable,du2021neural,gafni2021dynamic,gao2021dynamic,hedman2021snerg,jiang2022neuman,kwon2021neural,li2021neural,nguyen2022free,park2020deformable,peng2022selfnerf,peng2021animatable,peng2020neural,pumarola2020d,raj2021pva,tretschk2021non}.

\noindent \textbf{Static Scene Representations}. 
SRN~\cite{sitzmann2019scene} represents scenes as continuous functions that maps 3D coordinates to a feature representation of local scene properties and formulates the image as a differentiable ray-marching algorithm.
NSVF~\cite{liu2020nsvf} utilizes a sparse voxel octree to represent a set of voxel-bounded
implicit fields. A differentiable ray-marching operation is adopted to render views from a set of posed RGB images.
NeRF~\cite{mildenhall2020nerf} optimizes a neural radiance field for a scene, which maps 3D coordinates and viewing directions to density and color using a neural network.  
While NeRF can render photo-realistic images given dense images as input, it is limited mostly to static scenes. 

\noindent \textbf{Dynamic Scene Representations}. Dynamic NeRFs~\cite{pumarola2020d, peng2020neural} extend NeRF to dynamic scenes by introducing a latent deformation field or human poses.
NeuralBody~\cite{peng2020neural} proposes a set
of latent codes shared across all frames anchored to a human body model in order to replay character motions from arbitrary view points under training poses. 
HumanNeRF~\cite{weng2022humannerf} or A-NeRF~\cite{su2021nerf} learn the motion information by combining the skeletal and the non-rigid transformations.
For these methods, the synthesis fails under novel poses. 
Human pose based representation can model the body shape at any time step but will fail to capture detailed appearance. 
To overcome this problem, we propose to construct the appearance-based representation by utilizing the 2D features anchored to the point clouds as an input.

\noindent \textbf{Dynamic Scene Fusion}.
To model the temporal cues across frames, previous works~\cite{dou2016fusion4d,kwon2021neural,li2021neural,xu2021h,yu2018doublefusion} combine motion information and introduce animatable avatar approaches~\cite{gao2022mps,geng2023learning,jiang2022instantavatar,li2022tava,liu2021neural,mihajlovic2022keypointnerf,su2021nerf,te2022neural,wang2022arah,weng2022humannerf,xu2022surface,zhao2022humannerf,zheng2022structured}. Some of these approaches rely on keypoints~\cite{mihajlovic2022keypointnerf}, correspondences~\cite{li2022tava} or vertex normal alignment~\cite{xu2022surface} to generate details. 
Li \etal~\cite{li2021neural} learn dense scene flow fields that learn 3D offset vectors from a point in time $t$ to the same point in time $t$-1 and time $t$+1. 
The offsets are implicitly supervised with 2D optical flow. 
Kwon \etal~\cite{kwon2021neural} employ a temporal transformer to integrate skeletal features across different frames. 
The vertices of the human body are first reprojected to the 2D plane and then image features are sampled to obtain the skeletal features.
Although both \cite{kwon2021neural} and our method use a temporal transformer, the way we use the transformer is substantially different. Kwon \etal~\cite{kwon2021neural} use the transformer to combine pixel-aligned skeletal features obtained by projecting the vertices to the 2D image plane and then sampled from images using bilinear interpolation. 
They require multiple views as input as inaccurate features are extracted when projecting the 3D vertices into single view.
Instead of combining skeletal features, we propose to use a transformer to combine the pose codes for any 3D point and its tracked points. Our method optimizes the 3D feature volumes and does not require multi-view inputs. 
\setlength{\abovedisplayskip}{0.15cm}
\section{Methodology}
\vspace{-0.15cm}

Given a monocular video of a human in motion, we synthesize free-viewpoint videos of the person under novel views and new poses. 
During training multi-view videos are utilized to train our pipeline. 
We denote the set of input video frames as $\{I_t|t=1,...,N_f\}$, where $t$ represents the frame index and $N_f$ is the number of frames. 
To avoid the influence of background changes due to the camera movement, we remove the background color with the mask using~\cite{lin2022robust} and only focus on the human in the foreground. The overview of our approach is illustrated in Figure~\ref{fig:pipeline}.

\vspace{-0.1cm}
\subsection{Neural Radiance Fields} \label{nerf_formula}
\vspace{-0.15cm}
NeRF~\cite{mildenhall2020nerf} represents a static scene as a radiance field and renders color using volume rendering~\cite{kajiya1984ray}. 
It utilizes the 3D location $\bm{x}=(x,y,z)$ and 2D viewing direction $\bm{d}$ as input and outputs color $\bm{c}$ and volume density $\sigma$ with a network for any 3D point:
\begin{equation}
F_{\Theta}:(\gamma_x({\bm x}), \gamma_d({\bm d})) \rightarrow (\bm{c}, \sigma).
\end{equation}
$\gamma_x$ and $\gamma_d$ are the positional encoding functions for viewing direction and spatial location, respectively. To render the pixel color, NeRF uses the volume rendering integral equation by accumulating volume densities and colors for all sampled points along the ray. 
Let $\bm{r}$ be the camera ray emitted from the center of projection to a pixel on the image. 
The pixel color bounded by $h_n$ and $h_f$ is given by:
\begin{align}
    \Tilde{C}(\bm r) = \int_{h_n}^{h_f} T(h) \sigma(\bm{r}(h))\bm{c}(\bm{r}(h),\bm{d})dh,
    \label{eq:nerf_equa}
\end{align}
where $T(h) = \exp (- \int_{h_n}^h \sigma(\bm{r}(s)) ds)$ denotes the accumulated transmittance along the ray from $h_n$ to $h$.
NeRF is scene-specific with known camera parameters, and renders photo-realistic scenes with. 
To extend NeRF to model dynamic humans, we propose to learn the implicit representation to represent the shape and appearance information of the human. 
Specifically, we introduce a pose-conditioned representation \emph{shared by all frames} and an appearance-conditioned representation \emph{specific to each frame}.

\vspace{-0.1cm}
\subsection{Pose-conditioned Representation}\label{pose_representation}
\vspace{-0.1cm}
Following~\cite{peng2021animatable,peng2020neural}, we assume the 3D human model is given for each frame (\ie we use the pre-computed available SMPL body fits or do the body fitting at each frame as a pre-processing step). 
We first extract the vertices from the posed 3D mesh and aim to learn a set of pose codes $\bm Z=\{\bm z^1, \bm z^2, \ldots,\bm z^{N_m}\}$ anchored to the vertices of the human body model. Here $N_m$ denotes the number of codes whereas the dimension of each pose code is set to 16 similar to~\cite{peng2020neural}. 
The implicit representation is then learned by forwarding the pose code into a neural network, which represents the geometry and shape of a human performer. 
The pose space is shared across all frames, which can be treated as a common canonical space and enables the representation of a dynamic human based on the NeRF. 
Finally a neural network learns the density and color for any 3D point and volume rendering is used to render per-pixel RGB values. 

The pose codes anchored to the body model are relatively sparse in the 3D space and as such, directly calculating the pose codes using trilinear interpolation would lead to less effective features for most points.
During our experimental investigation we identified that a SparseConvNet is the right design choice as it propagates the codes defined on the mesh surface to the nearby 3D space. The SparseConvNet encodes the pose codes with $N_m$ vertices which correspond to $N_m$ pose codes are optimized during training.
To acquire the pose code for each point sampled along the camera ray, we use trilinear interpolation to query the code at continuous 3D locations. 
Here the pose code for the $i$-th point $\bm x_t^i$ at frame $t$ is represented by $\phi(\bm x_t^i,\bm Z)$ and is then fed to a neural network to predict the density and color.
The pose codes learned in the shared space of all frames model the human shape well in both known and unseen poses. 
However, the synthesized views still lack details under novel poses without optimizing the pose codes. Hence to model per-frame details, we propose an appearance-conditioned implicit representation using the monocular image and its predicted depth as the reference inputs.
\vspace{-0.1cm}
\subsection{Appearance-conditioned Representation}\label{appear_representation}
\vspace{-0.1cm}
An image along with its predicted depth can serve as the appearance human body prior under a single view. 
To learn detailed information at each individual frame, we learn the appearance code anchored to the point clouds.
The point clouds are obtained by lifting the RGB to the 3D space using the depth image which is generated by finetuning a state-of-the-art depth estimation method~\cite{yuan2022new} on our dataset. In that way the point clouds model the partially visible body of the human performer and capture details such cloth wrinkles. 
Given a 2D pixel $\bm{p}_t^i$ and its corresponding depth value $d_t^i$, the point cloud generation process is formulated as 
\(\bm{p}^{i}_t = F(\bm{p}_t^i, d_t^i)\)
where $\bm{p}^{i}_t$ is the generated 3D point for frame $t$, and $F(\cdot,\cdot)$ is the function generating a 3D point given a 2D pixel and a camera pose $\{\mathbf{K}_t, \left[\mathbf{R|t}\right]_t\}$.
Different from the pose-conditioned latent codes that are shared across all frames, the proposed appearance-conditioned codes are anchored to the point clouds, which are obtained from the pixel-aligned features extracted from the image encoder $E$. %
To take advantage of the rich semantic and detailed cues from images, we use ResNet34~\cite{he2016deep} to encode the image feature map $E(I_t)$ for the input image $I_t$.
Specifically, we first extract features from the ResNet34, that are passed to three Conv2D layers that reduce the dimension followed by a SparseConvNet to encode the features anchored to the sparse point clouds. 
To obtain the appearance code for each point sampled along the camera ray, we use trilinear interpolation to query the code at the continuous 3D locations. 
$\psi(\bm{x}_t^i,E)$ is adopted to represent the appearance code for point $\bm{x}_t^i$. 
The appearance code together with the pose code are forwarded into a neural network to predict the density and color.
The appearance code learned on each single frame models the details on the human body and recovers some missing pixels in the 3D space.

\subsection{Temporal Fusion Module}\label{transformer}
\vspace{-0.1cm}
Frames from different timesteps provide complementary information to the query frame, and will be referred as key frames. A temporal transformer then effectively integrates the features extracted from the query and key frames.
To obtain the corresponding pixels in the key frame, we use the parametric body model of each frame to track the points. 

\noindent\textbf{Point Tracking}. First, $N_a$ points on each face of the mesh are randomly sampled, which results in $N_s \times N_a$ points on the whole body surface where $N_s$ represents the number of faces.
We calculate the distance between a 3D point sampled on the camera ray and all points on the surface at the query frame $I_t$. 
We keep each sample $\bm{x}_t^i$ close to the
surface for rendering the color if $\min_{v\in \mathcal{V}_t}||\bm{x}_t^i-v||_2<\gamma$ and obtain the nearest point $\bm{\hat{x}}_t^i$ on the surface at frame $I_t$, where $\mathcal{V}_t$ is the set of sampled points. 
Furthermore, we track the points at different frames that match $\bm{\hat{x}}_t^i$ by the body motion, and assign the feature of the tracked points to $\bm{x}_t^i$.

\noindent\textbf{Key Frame Selection}. We automatically select three key frames from training frames. 
We first rotate the human pose along the Y-axis by 90$^{\circ}$, 180$^{\circ}$ and 270$^{\circ}$ and calculate the distances between all training poses and the rotated poses for the query frame $\bm{S}_t$ by $||\bm{S}_t-\bm{S}_j||_2$. We keep the frames with the $K$-NN distances, where $j$ is the index of the training frames, $\bm{S}$ are the coordinates of the vertices extracted from the body mesh and $K$ is set to 1 for each rotated frame.

\noindent\textbf{Temporal Fusion}. After obtaining the pose codes 
from $N$ frames ($K_f$ key frames and one query frame), a transformer based structure~\cite{dosovitskiy2020image} is introduced that takes the $N$ features as input and employs a multi-head attention mechanism along with an MLP for feature aggregation.
The fusion module is described in Figure~\ref{fig:pipeline}.
Query pose code $\phi(\bm x_{t}^i, \bm Z)$ is combined with the key frame pose codes by using the attention weight.
Here we use $f_q(\cdot), f_k(\cdot)$ and $f_v(\cdot)$ generated by fully-connected layers to represent the query, key and value.
The query and the key are used to calculate the attention map using the multiplication operation, which represents the correlation between the query pose code and the key pose code. 
The attention map retrieves all key pose codes and combines with the value by an addition operation.
Formally, the attention weight for point $\bm x_t^i$ in frame $t$ and tracked point $\bm x_k^i$ in frame $k$ is calculated by:
\begin{equation}
    a_{t,k}^i = \Omega\left(\frac{f_q(\phi(\bm x_t^i, \bm Z)) \cdot f_k(\phi(\bm x_k^i, \bm Z))^{\top}}{\sqrt{d}}\right),
\end{equation}
where $\sqrt{d}$ is a scaling factor based on the network depth, and $\Omega(\cdot)$ denotes the softmax operation. 
The aggregated feature for input $\phi(\bm x_t^i, \bm Z)$ is formulated as:

\begin{equation}
   \phi'(\bm x_t^i, \bm Z) = \sum_{k\in \mathcal{K}}\phi(\bm x_t^i, \bm Z) \cdot a_{t,k}^i + f_v(\phi(\bm x_t^i, \bm Z)),
\end{equation}
where $\mathcal{K}$ denotes the index set of the combined frames.
In this work, multi-head self-attention is adopted by running multiple self-attention operations, in parallel. The results from different heads are integrated to obtain the final output.
After the self-attention mechanism, each input feature contains its original information and also takes into account the information from all other frames. 
As such, the information from key frames and the query frame are combined together. 
Average pooling is then employed to integrate all features, which serves as the output of the temporal fusion module. 
In our implementation, we do not adopt any positional encoding on the input feature sequence.

\subsection{Density and Color Regression}\label{volume_rendering}
Figure~\ref{fig:pipeline} shows the prediction of density and color that are represented by a neural network. 
For each frame, the network takes the pose code, appearance code, spatial location and viewing direction as the input and outputs the density and color for each point in the 3D space.
Similar to~\cite{mildenhall2020nerf, rahaman2019spectral}, we apply positional encoding to both the viewing direction $\bm d$ and the spatial location $\bm x$ by mapping the inputs to a higher dimensional space. 
For frame $t$, the volume density and color at point $\bm x_t^i$ is predicted as a function of the latent codes, which is defined as:
\begin{equation}
(\sigma_t^i, \bm c_t^i) = M(\phi(\bm x_t^i,\bm Z),\psi(\bm x_t^i,E), \gamma_d(\bm d_t^i), \gamma_x(\bm x_t^i)), 
\end{equation}
where $M(\cdot)$ represents a neural network. 
$ \gamma_d(\bm d_t^i)$  and $ \gamma_x(\bm x_t^i)$ are the positional encoding functions for viewing direction and spatial location, respectively. 

\subsection{Objective Functions}\label{loss_function}
The objective function of our approach is defined as $\mathcal{L}=\mathcal{L}_{c1}+\mathcal{L}_{c2},$
where $\mathcal{L}_{c1}$ is the reconstruction loss for the rendered pixels and $\mathcal{L}_{c2}$ is the image loss for the image decoder network $D$.
The image decoder comprises multiple Conv2D layers behind the ResNet34, and aims to reconstruct the input image. The reconstruction loss forces the encoder to be optimized and generates better pixel-aligned features.
We render the color of each ray using both the coarse and fine set of samples, and minimize the mean squared error between the rendered pixel color $\Tilde{C}_c(\mathbf{r})$ and ground-truth color $C(\mathbf{r})$ for training:

\begin{equation}
\mathcal{L}_{c1}=\sum_{\mathbf{r}\in R}||\Tilde{C}_c(\mathbf{r}) - C(\mathbf{r})||_2^2 + ||\Tilde{C}_f(\mathbf{r}) - C(\mathbf{r})||_2^2,  
\end{equation}
where $R$ is the set of rays.  $\Tilde{C}_c$ and $\Tilde{C}_f$ denote the predictions of the coarse and fine networks. Finally, $\Tilde{I}(\bm p)$ and ${I}(\bm p)$ are the reconstructed and ground truth colors for pixel $\bm p$ in the set of pixels $\mathcal{I}$ and are used to compute the image loss:

\begin{equation}
\mathcal{L}_{c2}=\sum_{\bm p\in \mathcal{I}}||\Tilde{I}(\bm p) - I(\bm p)||_2^2.
\end{equation}

\begin{table*}[t]
\setlength{\tabcolsep}{0.4em}
\begin{center}
\small
\caption{\textbf{Quantitative Comparison} on training (top 5 rows) and novels views (bottom 5 rows) under novel poses across all sequences.}
\scalebox{0.91}{
\begin{tabular}{lccccccccccccc}
    \toprule
    & &\multicolumn{2}{c}{\emph{Sequence1}}  &  \multicolumn{2}{c}{\emph{Sequence2}} & \multicolumn{2}{c}{\emph{Sequence3}} & \multicolumn{2}{c}{\emph{Sequence4}} &
    \multicolumn{2}{c}{\emph{Sequence5}} \\
    \midrule
    Models & Train Views & PSNR$\uparrow$ & SSIM$\uparrow$& PSNR$\uparrow$  & SSIM$\uparrow$ &PSNR$\uparrow$ & SSIM$\uparrow$ &PSNR$\uparrow$ & SSIM$\uparrow$
    &PSNR$\uparrow$ & SSIM$\uparrow$\\
    \midrule
    NeuralBody & \checkmark &22.83&0.79&14.12&0.46&\textbf{19.08}&\textbf{0.77}&23.84&0.73 & 24.66&0.83\\
    NHP & \checkmark &22.16&0.75&13.28&0.43&18.89&0.76&23.45&0.75 & 24.34&0.80\\
    Ani-NeRF & \checkmark &22.03&0.74&13.26&0.40&18.65&0.69&23.33&0.68 & 24.21&0.79\\
     HumanNeRF & \checkmark &23.52&0.80&14.36&0.47&18.97&0.79&24.62&0.74&24.59&0.82\\
    \textbf{Ours} & \checkmark & \textbf{24.76} & \textbf{0.81} & \textbf{15.51} & \textbf{0.49}&18.78 & 0.72 & \textbf{24.84}& \textbf{0.76} &\textbf{24.78}  & \textbf{0.83}\\
    \midrule
    NeuralBody &  & 22.76 &0.79  & 13.52 &0.41 &\textbf{19.91} & \textbf{0.79} &23.81  &\textbf{0.77}  & 23.17  & 0.78\\
    NHP &  & 21.96 &0.77  & 13.18 &0.41&19.67 & 0.77  &23.42  &0.73  & 22.76  & 0.74\\
    Ani-NeRF  &  & 21.86 &0.71  & 13.03 &0.37&18.32 & 0.64  &22.67 &0.66  & 22.68  & 0.75\\
    HumanNeRF  &  &23.13&0.79&13.93&0.42&19.88&0.76&23.89&0.76&23.83&0.79\\
    \textbf{Ours} &&\textbf{24.67} & \textbf{0.80} & \textbf{15.01} &\textbf{0.45}&18.70&0.71 & \textbf{24.63} & 0.75 &\textbf{23.91}  &\textbf{0.81} \\
    \bottomrule
\end{tabular}}
\label{tab:quan_novel_view}
\end{center}
\vspace{-8mm}
\end{table*}

\vspace{-0.4cm}
\section{Implementation Details}

\noindent \textbf{Network Details}. For the encoder $E$, we extract a feature pyramid~\cite{lin2017feature} from each image similar to~\cite{yu2021pixelnerf}.
A ResNet34 backbone pretrained on ImageNet is utilized for our experiments.  
The output feature of the decoder has $1/4$ the spatial resolution compared to the input image. 
Multi-scale features are extracted prior to the fourth pooling layer. 
We extract pixel-aligned features using bilinear interpolation, and then concatenate them to form a latent vector of size 256. 
To construct the image decoder $D$, we simply connect several Conv2D/Upsampling layers to reconstruct the input image.

For the depth prediction network, we utilize~\cite{yuan2022new} which takes a single frame as input and outputs a depth map. This network is fine-tuned using our training images and ground truth depths. For inference, the depth map is predicted and is then processed by lifting the pixels to the 3D space to remove the points outside the dilated posed mesh. 
For the videos without ground truth depths for training, we use the depth map predicted by the NeuralBody~\cite{peng2020neural}.

For the transformer network, we utilize three heads for the self-attention module, which has a similar structure as~\cite{dosovitskiy2020image}. 
Following NeRF~\cite{mildenhall2020nerf}, we perform hierarchical volume sampling and simultaneously optimize a coarse and fine network with identical network architecture. %
At the coarse scale, we sample a set of $M_c$ points using stratified sampling. %
With the prediction of the coarse network, we then sample another set of points along each camera ray, where samples are more likely to be located at relevant regions for rendering. 
We sample additional $M_f$ locations and use all $M_c+M_f$ locations to render the fine results, where $M_c$ and $M_f$ are set to 64.

\noindent \textbf{Training Details}. We train all layers using Adam~\cite{kingma2014adam} with base learning rates for the encoder-decoder network and other layers set to $10^{-3}$ and $5\times10^{-4}$, which decay exponentially during the optimization. Additional network architecture as well as implementation details are provided in the supplementary material. 
It takes about 48 hours using 4 GeForce RTX 3090 GPUs to train our method for 200 frames and 30 views each. The inference time on a single image is \(\mytilde50s\). Note that fast training is not a primary goal of our work and several recent techniques~\cite{chen2022tensorf, muller2022instant} for accelerating our training regime could be used in the future. 

\section{Experiments}

\noindent \textbf{Datasets}.
To train our method, we rely on the \emph{proposed dataset} (four sequences of real humans in motion that captured with a 3dMD full-body scanner and a single sequence of a synthetic human in motion) and the public \emph{ZJU-MoCap dataset}~\cite{peng2020neural}. 
The 3dMD body scanner comprises 18 calibrated RGB cameras that capture a human in motion performing various actions and facial expressions and output a reconstructed 3D geometry and texture per frame. 
These scans are noisy but capture facial expressions and fine-level details like cloth wrinkles.
The synthetic scan is a high-res animated 3D human model with synthetic clothes (T-shirt and pants) that were simulated. 
Unlike the 3dMD scans, the 3D geometry is clean but lacks facial expressions.
We render RGB and Depth for all sequences from 30 views that cover the whole hemisphere (similarly to the way NeRF data are generated) at 6 fps using Blender Cycles. 

Each video has more than 200 frames of 1024$\times$1024 resolution.
For the proposed dataset, we select the first half of the frames for training and the rest for inference. 
For the ZJU-MoCap dataset, we use the same training frames as~\cite{peng2020neural}.
Both training and test frames contain large variations in terms of the motion and facial expressions. 
At training and testing, a single image at each frame is used as the input.
All the input images at different frames share the same static camera pose. In addition, 29 (proposed dataset) or 14 (ZJU-MoCap) more views with different camera poses are used to train the network. The output is a rendered view given any camera pose (not including the camera pose of the input image).

\begin{figure*}[t]
    \centering
    \includegraphics[width=1.\linewidth]{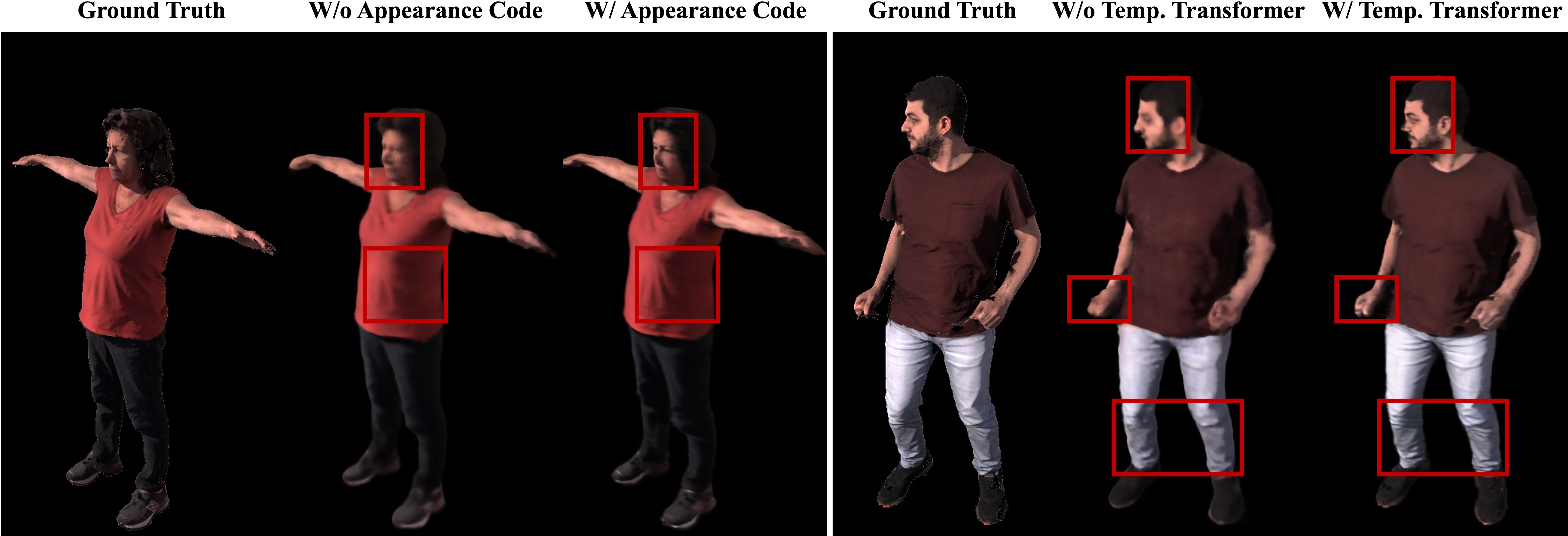}
    \caption{Impact of the appearance code and temporal transformer on synthesized images. Both components help recover better details.}
    \label{fig:applications_appearance_trans}
    \vspace{-3mm}
\end{figure*}

\noindent \textbf{Baselines for Quantitative Comparisons}.

\begin{compactitem}
    \item NeuralBody~\cite{peng2020neural} models dynamic scenes using latent codes anchored to the human pose as an extra input besides the coordinate and viewing direction.
    \item NHP~\cite{kwon2021neural} extends NeuralBody to a generalizable model by aggregating temporal and multi-view pixel-aligned features. 
    We remove the multi-view branch with only a monocular video as the input.
    \item Ani-NeRF~\cite{peng2021animatable}  combines NeRF and 3D human skeletons by learning the blend weight field to recover animatable human models.
    \item HumanNeRF~\cite{weng2022humannerf} learns a volumetric representation of the
    person in a canonical T-pose and a motion field that maps the estimated canonical representation to every frame of the video.
\end{compactitem}

\noindent \textbf{Evaluation Metrics}.
Following existing approaches~\cite{mildenhall2020nerf, peng2020neural}, we evaluate the performance on the proposed dataset using two metrics, including the peak signal-to-noise ratio (PSNR) and structural similarity index (SSIM).

\subsection{Experimental Results}
We conduct a wide range of quantitative and qualitative comparisons to demonstrate the key contributions of our work along with ablation studies against simplified variants where proposed modules are removed. We evaluate our approach against the aforementioned works on the task of view synthesis for \emph{new unseen poses}. All methods except HumanNeRF (monocular video for training) use the same camera views and human poses during the training and test stages for fair comparison.

\noindent \textbf{Novel View and Novel Pose}. We quantitatively evaluate our approach on novel views of all sequences and report our findings at the bottom part of Table~\ref{tab:quan_novel_view} and Table~\ref{tab:zjumocap}. 
The proposed method clearly outperforms other competitive approaches on most video sequences. 
All other methods, have a hard time generating realistic renders when the test poses deviate significantly from what was seen during training. 
On the contrary, our proposed method generates fine-level details on the human body, which indicates that the point clouds-based representation and the temporal transformer can help reason the missing structure and recover the non-visible parts at the unseen poses.

\noindent \textbf{Training View and Novel Pose}.
The quantitative results on training views and novel poses are shown at the top part of Table~\ref{tab:quan_novel_view} and Table~\ref{tab:zjumocap}. We conduct this experiment to showcase that our approach generalizes well to unseen poses as the difference between training and testing views is fairly small. 
On most video sequences, the proposed method performs better compared to all other baselines. 

\noindent \textbf{ZJU-MoCap Results}. In Table~\ref{tab:zjumocap} we provide quantitative comparisons on the publicly available ZJU-MoCap dataset and show that clearly outperforms all prior work on both training and novel views under unseen poses. Our method benefits from the intra-frame point clouds representation and inter-frame temporal information.

\begin{table}[t]
\setlength{\tabcolsep}{0.4em}
\begin{center}
\small
\caption{\textbf{ZJU-MoCap - Quantitative Comparison} on training (top 4 rows) and novels views (bottom 4 rows) under novel poses on \emph{Subject 313} and \emph{Subject 393}. See Supplemental for more.}
\scalebox{0.91}{
\begin{tabular}{lccccccccccccc}
    \toprule
    & &\multicolumn{2}{c}{\emph{Subject 313}}  &  \multicolumn{2}{c}{\emph{Subject 393}}  \\
    \midrule
    Models & Train Views & PSNR$\uparrow$ & SSIM$\uparrow$& PSNR$\uparrow$  & SSIM$\uparrow$\\
    \midrule
    NeuralBody & \checkmark &23.86&0.88&22.78&0.86\\
    Ani-NeRF & \checkmark &23.94&0.88&23.01&0.87\\
    HumanNeRF & \checkmark &22.61&0.83&21.84&0.84\\
    \textbf{Ours} & \checkmark &\textbf{24.41}&\textbf{0.89}&\textbf{23.63}&\textbf{0.87}\\
    \midrule
    NeuralBody &&23.70&0.87&22.53&0.84\\
    Ani-NeRF  &&23.47&0.86&22.39&0.85\\
    HumanNeRF  &&22.53&0.83&21.72&0.83\\
    \textbf{Ours} &&\textbf{23.93}&\textbf{0.88}&\textbf{22.90}&\textbf{0.86} \\
    \bottomrule
\end{tabular}}
\vspace{-5pt}
\label{tab:zjumocap}
\end{center}
\vspace{-6mm}
\end{table}

\noindent \textbf{Qualitative Results}. We qualitatively compare our approach other top-performing methods under novel poses in Fig.~\ref{fig:applications_vis}. 
With the human pose as the geometric guidance, NeuralBody predicts the body shape well but fails to render fine-level details on the human body.
NeuralBody does not generalize well to novel poses because the shared latent codes across all frames are not optimized during the test stage.  
Ani-NeRF and HumanNeRF use skeleton as input, which ends up overfitting to the training poses and generates blurry results for unseen poses. The qualitative comparisons are provided in Fig~\ref{fig:applications_vis} where our approach captures fine-level details on the body (\(1^{st}, 3^{rd}\) rows) and head (\(2^{nd}\) row) better than prior works~\cite{peng2021animatable,peng2020neural,weng2022humannerf}.

\noindent{\textbf{Limitations}.
The temporal transformer recovers more non-visible pixels in the body. Without encoding facial expressions, our method can handle humans without substantial expression variations. 
However when the the query frame has a facial expression different from the key frames our method predicts blurred facial expressions as combining the key frames with the query frame makes the network unable to differentiate the specific facial characteristics in the query frame. 
Future work will encode facial expressions for each frame as a separate code and thus being able to render such diverse expressions under new views. 

\begin{figure*}[t]
    \centering
    \includegraphics[width=\linewidth]{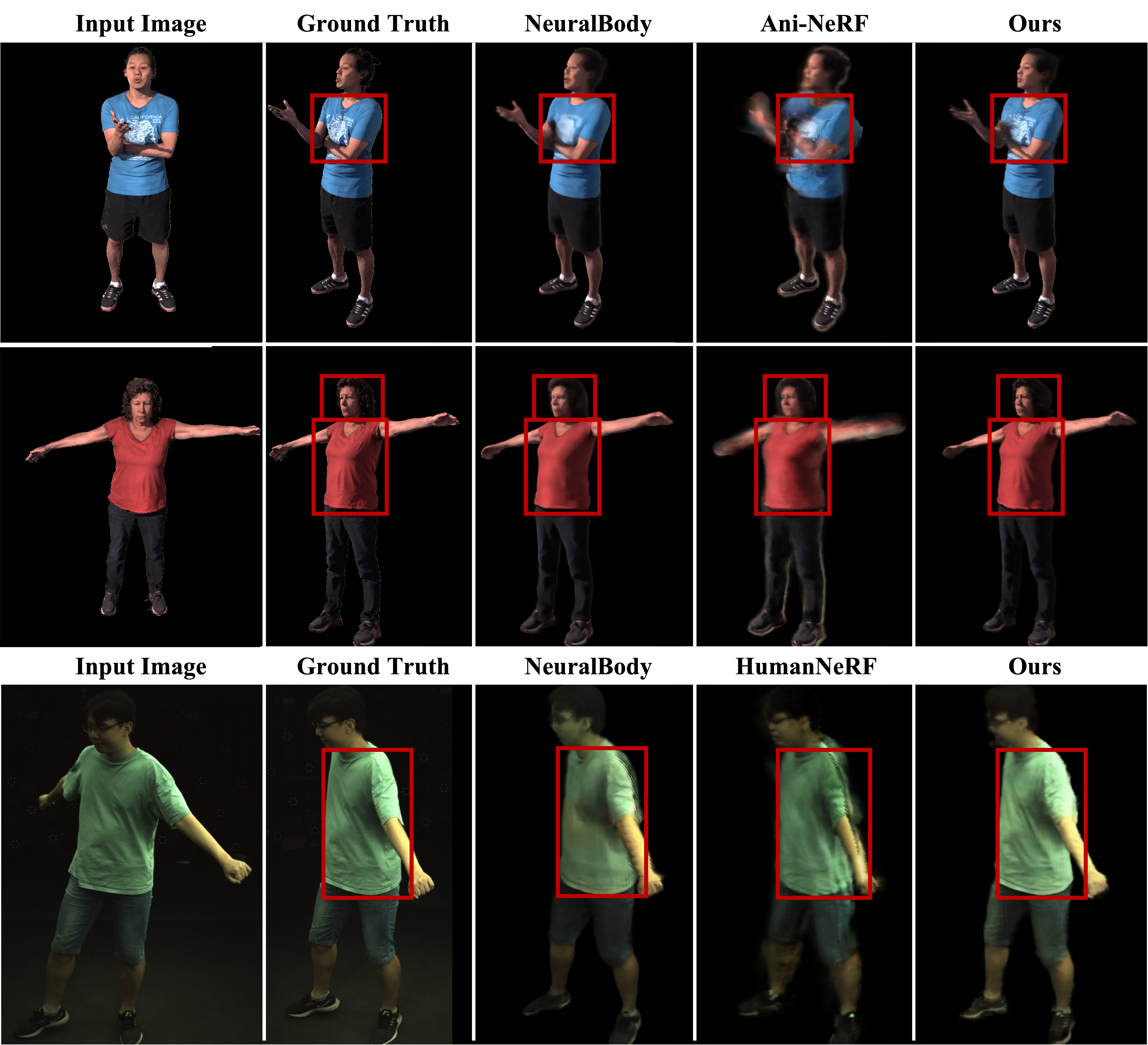}
    \caption{\textbf{Qualitative Results.} Our proposed approach achieves significantly better render quality compared to NeuralBody, Ani-NeRF and HumanNeRF across various poses, viewpoints and subjects. All poses are unseen and have not been used for training.}
    \label{fig:applications_vis}
    \vspace{-4mm}
\end{figure*}

\subsection{Ablation Studies}

\setlength{\tabcolsep}{.04cm}
\begin{table}[t]
	\centering
	\footnotesize
	\caption{\textit{\textbf{Left}}: Ablation Study of appearance-based representation and transformer fusion module on \emph{Sequence1}. \textit{\textbf{Right}}: Effect of the number of key frames on \emph{Sequence1}.}
	\begin{tabular}{lcc}
    \toprule
    Model Variant & PSNR$\uparrow$ & SSIM$\uparrow$\\
    \midrule
    W/o appearance-based repr. &23.65&0.79 \\
    W/o temporal transformer
    &23.70&0.77\\
    W/o both &22.78&0.75\\
    \textbf{Ours}  &\textbf{24.76}   &\textbf{0.81}     \\
    \bottomrule
\end{tabular}
    \;\;\;\ 
    \begin{tabular}{lcc}
    \toprule
    \# frames & PSNR$\uparrow$ & SSIM$\uparrow$\\
    \midrule
    1 &24.17&0.80 \\
    3 (Ours)
    &24.76   &0.81\\
    5  &24.84   &0.81     \\
    \bottomrule
    \end{tabular}
    \label{tab:aba_appearance_code}
    \vspace{-4mm}
\end{table}

\noindent{\textbf{Effect of the Appearance-based Representation.}} 
Using the appearance code brings performance improvements (Table~\ref{tab:aba_appearance_code} and Fig.~\ref{fig:applications_appearance_trans}) on the fine structures (cloth wrinkles, facial expressions)  
in different parts of the body, which demonstrates that the appearance code anchored to the points clouds helps recover the missing pixels in the query view.

\noindent{\textbf{Effect of the Temporal Transformer}}. As shown in Table~\ref{tab:aba_appearance_code} and Fig.~\ref{fig:applications_appearance_trans}, temporal fusion module can help the model generate better rendering performance. We observe that the details like wrinkles on the pants are finer, the hands are cleaner and the face is significantly more crisp. 

\noindent{\textbf{Effect of the Number of Key Frames}}. To evaluate the impact of the number of key frames, we report the performance in Table~\ref{tab:aba_appearance_code}. We observe that the performance increases with more key frames and saturates with 5 frames. 

\noindent{\textbf{Effect of the Depth Estimation}}. Our proposed approach relies on depth estimation that allows us to lift the RGB input to 3D and obtain the pointcloud that is then fed to our 3D backbone architecture. To identify the impact of the depth estimation module we conducted an ablation study where ground-truth depth is utilized for \emph{Sequence 4} of the proposed dataset. We observed that PSNR and SSIM are 24.96 and 0.77 (compared to 24.84 and 0.76) respectively when using ground-truth depth which are just slightly higher than using our depth estimation model. This indicates that even using the inaccurate depth information, our method can generalize well to the unseen poses.
For videos without ground truth depth (\ie ZJU-MoCap data) we use the depth predicted by NeuralBody and despite relying on their estimation, we clearly outperform them (shown in Tab.~\ref{tab:zjumocap}) which showcases the importance of using depth information for the rendering of articulated avatars.
\section{Conclusion}
\vspace{-2mm}
In this paper, we built upon recent advances of neural radiance fields pertaining to digital humans and addressed key challenges that existing human body based methods suffer from, preventing them to generalize well to unseen poses. Towards that direction, we proposed to integrate a pose code and an appearance code to synthesize humans in novel views and different poses with high fidelity.
The pose code that is anchored to the human pose models the human shape, whereas the appearance code anchored to the point clouds infers the fine-level details and recovers the missing parts. The point clouds are generated by lifting the 2D information to the 3D space using an estimated depth map.
To leverage temporal information, we proposed to use the body motion to track points from the query frame to a few automatically-selected key frames and adopted a temporal transformer to aggregate information across multiple frames. 
The transformer-based fusion module recovers the non-visible part in the query frame. 
Our approach achieves significantly better results against several prior methods under novel views and unseen poses with quality that has not been observed in prior work. We provided a plethora of experimental comparisons, qualitative results and ablation studies to back-up our claims and we showcase that fine-level information such as fingers, logos, cloth wrinkles and face details are faithfully rendered with high fidelity. 
\clearpage
{\small
\bibliographystyle{ieee_fullname}
\bibliography{References}
}

\end{document}